\documentclass[conference]{IEEEtran}
\IEEEoverridecommandlockouts

\usepackage{ulem}

\usepackage{cite}
\usepackage{amsmath,amssymb,amsfonts}
\usepackage{algorithmic}
\usepackage{graphicx}
\usepackage{textcomp}
\usepackage{xcolor}
\usepackage{booktabs}
\usepackage{multirow}
\usepackage{soul}
\usepackage{fontawesome}
\def\BibTeX{{\rm B\kern-.05em{\sc i\kern-.025em b}\kern-.08em
    T\kern-.1667em\lower.7ex\hbox{E}\kern-.125emX}}
\begin{document}

\title{Contrast-Unity for Partially-Supervised \\ Temporal Sentence Grounding
\vspace{-0.4cm}
}
\DeclareRobustCommand*{\IEEEauthorrefmark}[1]{%
  \raisebox{0pt}[0pt][0pt]{\textsuperscript{\footnotesize #1}}%
}
\author{
    \IEEEauthorblockN{
        Haicheng Wang\IEEEauthorrefmark{1,2,4}\textsuperscript{\textasteriskcentered}, 
        Chen Ju\IEEEauthorrefmark{2}\textsuperscript{\textasteriskcentered},
        Weixiong Lin\IEEEauthorrefmark{4}, 
        Chaofan Ma\IEEEauthorrefmark{4}, 
        Shuai Xiao\IEEEauthorrefmark{2},
        Ya Zhang\IEEEauthorrefmark{3}\textsuperscript{\faEnvelopeO},
        Yanfeng Wang\IEEEauthorrefmark{3}
    }
    \IEEEauthorblockA{
        \IEEEauthorrefmark{1}SJTU Paris Elite Insitute of Technology, Shanghai Jiao Tong University, China  \ 
        \IEEEauthorrefmark{2}Taobao \& Tmall Group of Alibaba, China \\
        \IEEEauthorrefmark{3}School of Artificial Intelligence, Shanghai Jiao Tong University, China
        \ 
        \IEEEauthorrefmark{4}CMIC, Shanghai Jiao Tong University, China\\
        Email: 
            \{anakin\_skywalker,wx\_lin, chaofanma, ya\_zhang, wangyanfeng622\}@sjtu.edu.cn, 
            cju.void@gmail.com
    }
}


\maketitle
\begingroup
\renewcommand\thefootnote{\relax}\footnotetext{*: Equal contribution. \quad \textsuperscript{\faEnvelopeO}: Corresponding author}
\endgroup
\begin{abstract}
Temporal sentence grounding aims to detect event timestamps described by the natural language query from given untrimmed videos. The existing fully-supervised setting achieves great results but requires expensive annotation costs; while the weakly-supervised setting adopts cheap labels but performs poorly. To pursue high performance with less annotation costs, this paper introduces an intermediate partially-supervised setting, \textit{i.e.}, only short-clip is available during training. To make full use of partial labels, we specially design one contrast-unity framework, with the two-stage goal of implicit-explicit progressive grounding. In the implicit stage, we align event-query representations at fine granularity using comprehensive quadruple contrastive learning: event-query gather, event-background separation, intra-cluster compactness and inter-cluster separability. Then, high-quality representations bring acceptable grounding pseudo-labels. In the explicit stage, to explicitly optimize grounding objectives, we train one fully-supervised model using obtained pseudo-labels for grounding refinement and denoising. Extensive experiments and thoroughly ablations on Charades-STA and ActivityNet Captions demonstrate the significance of partial supervision, as well as our superior performance.
\end{abstract}


\section{Introduction}
Temporal sentence grounding (TSG) plays an important role for video-language understanding, with the goal to detect the start and end timestamps of the event described by a given natural language query from untrimmed videos. 
TSG covers extensive application scenarios~\cite{b1,b2,b3}, as it could learn high-quality cross-modal representations from large-scale data.

TSG has developed two popular settings for data annotation: fully-supervised setting (FTSG)~\cite{b4,b5}, where each (video, query) pair is annotated with precise temporal boundaries, and weakly-supervised setting (WTSG)~\cite{b6,b7}, where only the corresponding (video, query) is provided without temporal annotations. While the fully-supervised approach is accurate, it is time-consuming and prone to subjective interpretation, especially for events with complex semantics. The weakly-supervised approach reduces annotation effort but results in lower performance, limiting its practical applications.

Hence, one question naturally arises: \textit{Is there an intermediate setting between full and weak supervisions in TSG, which can obtain relatively high performance but requires less annotation cost?}
This paper answers the question by introducing the \textbf{partially-supervised setting (PTSG)}. Specifically, for each text query, a partial temporal region corresponding to a short video-clip is annotated within the whole event interval. And in the strictest case, partial labels could degenerate to single-frame labels, \textit{i.e.}, labeling one timestamp for each event. At a slight more cost than WTSG in annotation time, such partial supervision greatly improves grounding performance, which is very effective comparing to full or weak supervisions.

Hereafter, our goal is to ground complete event intervals through limited yet precise partial labels. With the same data formulation (video, query, timestamps), partial supervision can approach full supervision continuously, by annotating a proper event duration. Thus, an intuitive thought is that, PTSG and FTSG can share the same training architecture. Following this idea, one trivial solution is to simply train FTSG model using partial annotations. As tested preliminarily, FTSG model performs well using high-quality partial annotation (80\% event coverage), proving its robustness for small turbulence. 
However, the limited short-clip partial label is too noisy for FTSG model to learn semantic patterns, resulting in an unsatisfying result. Therefore, We design \textbf{a contrast-unity framework} for implicit-explicit (two-stage) progressive grounding.

Given the training set with incomplete partial labels, \textbf{the implicit stage} aims to refine the partial annotation at fine granularity. 
To get better labels, we propose one novel quadruple contrast pipeline, leveraging inter and intra-sample contrast for uni and cross-modal alignment. 
The first two contrasts are built on intra-samples to promote event-query gather for cross-modal correspondence and raise event-background separation for visual uni-modality. Then, to build more semantic contrasts from the whole dataset, another two contrasts are proposed for inter-samples to further enable intra-cluster compactness and inter-cluster separability. To obtain refined event intervals, we introduce an event detector which takes partial labels as seed anchors and extends them for an event mask. Then, features for event and background can be calculated via the event mask.

Thanks to the essence of multi-instance learning~\cite{b8}, with well-alignment representations, the event detector can output refined grounding pseudo-labels. 
Next, we bridge \textbf{another explicit stage} after the implicit stage, by treating grounding pseudo-labels as ground-truth to train one fully-supervised model, then inference through this fully-supervised model. 
This framework could do more at one stroke. Structurally, it bridges the setting gap between PTSG and FTSG, enabling PTSG to enjoy advanced bonuses from FTSG, {\em e.g.}, superior architecture~\cite{b9,b10} and explicit grounding optimization. Functionally speaking, this framework is applicable from single-frame to fully-supervised TSG, giving more freedom to the annotation procedure. On two datasets: Charades-STA and ActivityNet Captions, we annotate partial labels, then experiment to reveal their significance. Our designed framework shows superior performance over competitors.

\section{Method}
\begin{figure}[t]
\begin{center}
\includegraphics[width=0.5\textwidth] {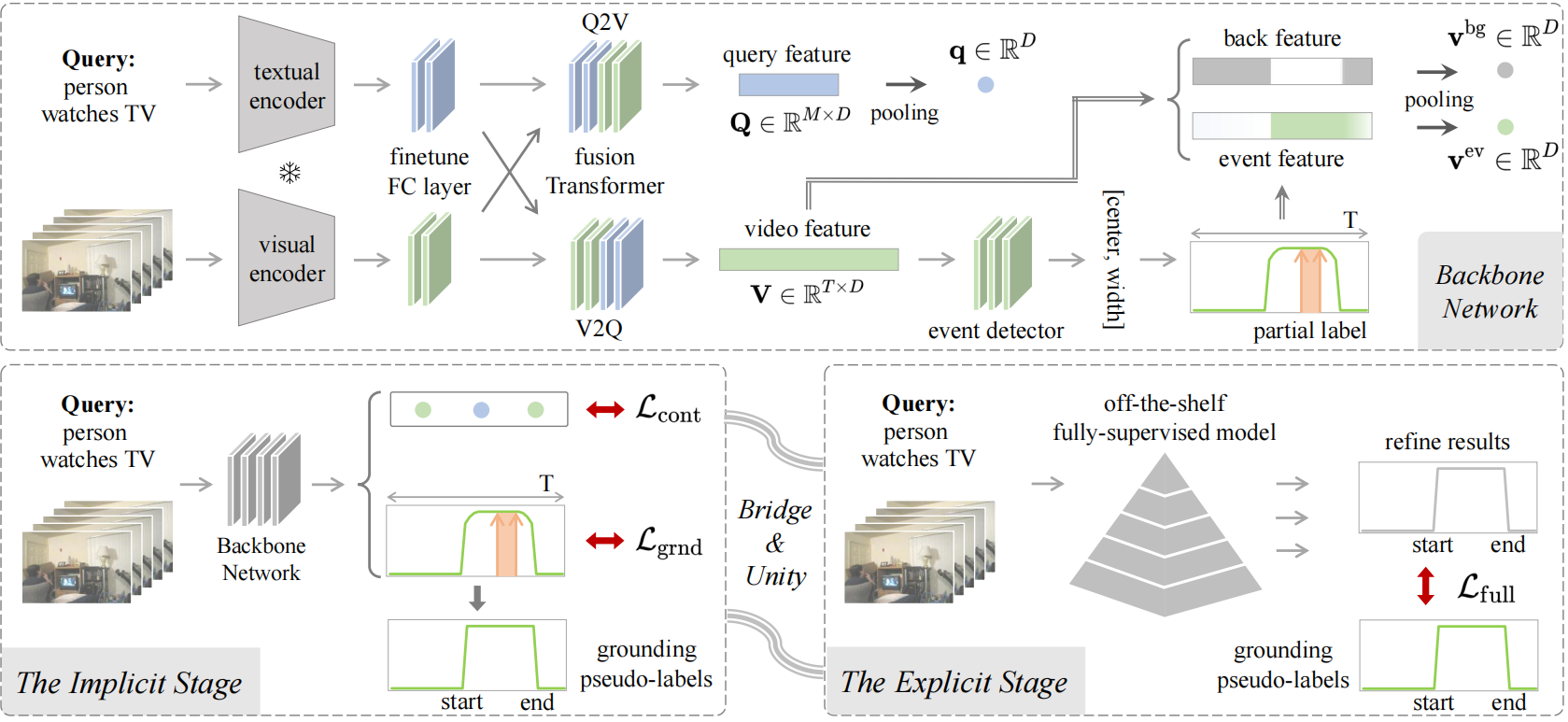}
\end{center}
\vspace{-0.4cm}
\caption{\textbf{Our Contrast-Unity Framework for PTSG.} It follows an implicit-explicit progressive pipeline. Given partial labels, the implicit stage makes fine-grained alignment of event-query representations, using quadruple contrastive learning. Such fine-aligned representations could naturally result in high-quality grounding psuedo-labels. Hereafter, the explicit stage takes these pseudo-labels as the ground-truth to train another fully-supervised model with explicit grounding objectives, for further grounding refinement. Such one framework achieves the unity between PTSG and FTSG.}
\vspace{-0.5cm}
\label{fig:framework}
\end{figure}

\subsection{Formulation \& Preliminaries}
\noindent \textbf{Problem Formulation.}  \label{subsec:Notations}
\hspace{0.1cm} Given an untrimmed video, Temporal Sentence Grounding (TSG) aims to detect the event boundary $(s, e) \in \mathbb{R}^2$ corresponding to given text query. Full supervision provides precise boundary $(s_i, e_i)$ for each query, while weak supervision only has video-query correspondence. To meet the demands of large-scale data annotations and strong performance, this work considers the novel \textit{partially-supervised TSG setting} (PTSG). For the $i$-th text query, only one short video clip $(t_i^s, t_i^e) \subseteq (s_i, e_i)$ is labeled. We could also write it as $(t_i^c, r)$, where $t_i^c$ is the clip center and $r$ is the clip range. In special case ($r=0$), this partial label degenerates to single-frame setting, \textit{i.e.}, only $t_i^c \in [s_i, e_i]$ is annotated.

\noindent \textbf{Feature Extraction \& Fusion.}
\hspace{0.1cm} We pre-extract features for videos and queries following the existing methods~\cite{b11,b12}. For the video stream, we adopt pre-trained 3D convolutional networks~\cite{b13,b14}, and obtain $\mathbf{V'} \in \mathbb{R}^{T \times D_v}$; for the query stream, we adopt GloVe~\cite{b15} to obtain $\mathbf{Q'} \in \mathbb{R}^{M \times D_q}$, where $T$, $M$, $D_v$, $D_q$ refer to the number of video frames, the number of query words, the video and query feature dimension. Then, we use one full-connection layer to individually fine-tune uni-modal features $\mathbf{V'}$ and $\mathbf{Q'}$ respectively. To interact bi-modal features, two cross-modal Transformers are further introduced. The fused visual features are denoted as $\mathbf{V} \in \mathbb{R}^{T \times D}$, and linguistic features as $\mathbf{Q} \in \mathbb{R}^{M \times D}$ ($D$ is the feature dimension).

\subsection{Implicit Stage: High-quality Representation}   \label{sec:implicit stage} 
\noindent \textbf{Event Detector.}
To perceive the time interval for event, we adopt a proposal-wise solution: treat $t^c$ as the seed anchor, and map video features $\mathbf{V}$ to center offset $\delta$ and event width $\ell$, through an event detector $\Phi(\cdot)$. And the corresponding start and end timestamps $[\widehat{s}, \, \widehat{e}]$ can be formulated as: 
\begin{equation} \label{eq:instance}
\setlength{\abovedisplayskip}{7pt}
    {[\delta, \; \ell] = \Phi(\mathbf{V}),  \quad  \widehat{s} = p-\frac{\ell}{2},  \quad  \widehat{e} = p+\frac{\ell}{2},}
\setlength{\belowdisplayskip}{7pt}
\end{equation}
where $p=t^c+\delta$ means the center of the grounded event.

\noindent \textbf{Event Representation.}
With the predicted start-end timestamps $[\widehat{s}, \, \widehat{e}]$, we first generate one differentiable temporal mask $\mathbf{m} \in \mathbb{R}^{T}$ through a learnable Plateau-shape~\cite{b16}; then filter out visual features for event $\mathbf{v}^{\mathrm{ev}}$, and background $\mathbf{v}^{\mathrm{bg}}$:
\begin{equation}
    {\mathbf{v}^{\mathrm{ev}} = \frac{1}{T}\sum_{t=1}^{T}{\mathbf{m_t}}\mathbf{V}_t,
    \ \ \
    \mathbf{v}^{\mathrm{bg}} = \frac{1}{T}\sum_{t=1}^{T}{(1-\mathbf{m_t})}\mathbf{V}_t.
    }
\end{equation}

\noindent \textbf{Quadruple Contrasts Pipeline.} To get refined pseudo-labels from event detector, we need to shape one high-quality visual-linguistic alignment space. We perform a quadruple-contrasts pipeline to pursue comprehensive alignment, covering intra- and inter-sample, uni- and multi-modality. The quadruple contrastive loss is calculated with a balancing parameter $\lambda$:
\begin{align}   \label{eq:cont}
\mathcal{L}_{\mathrm{cont}} = (\mathcal{L}_{\mathrm{raml}} + \mathcal{L}_{\mathrm{raun}})  + \lambda(\mathcal{L}_{\mathrm{erml}} + \mathcal{L}_{\mathrm{erun}}).
\end{align}

\noindent \textit{\uuline{Intra-Sample Contrastive Learning.}} 
We first consider learning from one single video-query sample, with visual feature for event $\mathbf{v}^{\mathrm{ev}}$, the background $\mathbf{v}^{\mathrm{bg}}$, and query feature $\mathbf{q} \in \mathbb{R}^{D}$ obtained by mean-pooling the word-wise feature $\mathbf{Q}$ available.

\noindent \underline{Event-Query Multi-Modal Contrast} enables event-query pairs gather in the embedding space. Here, we introduce the mean video feature $\mathbf{v}^{\mathrm{vd}}$ as one reference, to promote the semantic similarity of ($\mathbf{v}^{\mathrm{ev}}$, $\mathbf{q}$) to be greater than that of ($\mathbf{v}^{\mathrm{vd}}$, $\mathbf{q}$), since $\mathbf{v}^{\mathrm{vd}}$ contains both event and background.
\begin{align}
\mathcal{L}_{\mathrm{raml}} = \mathrm{max}(\mathcal{S}(\mathbf{v}^{\mathrm{vd}}, \mathbf{q})-\mathcal{S}(\mathbf{v}^{\mathrm{ev}}, \mathbf{q})+\alpha, 0),
\end{align}
where $\mathcal{S}$ and $\alpha$ are cosine similarity and margin parameter. $\mathbf{v}^{\mathrm{vd}}$ is got by pooling frame-wise video features $\mathbf{V}$.

\noindent \underline{Vision Uni-Modal Contrast.} 
Videos are fine-grained and continuous, resulting in similar features across event and background. We hence apply visual-modal contrastive learning to raise event-background separation. Similarly, we use triplet loss to distinguish (video-event) and (event-background).
\begin{align}
\mathcal{L}_{\mathrm{raun}} = \mathrm{max}(\mathcal{S}(\mathbf{v}^{\mathrm{ev}}, \mathbf{v}^{\mathrm{bg}})-\mathcal{S}(\mathbf{v}^{\mathrm{ev}}, \mathbf{v}^{\mathrm{vd}})+\beta, 0),
\end{align}

\noindent \textit{\uuline{Inter-Sample Contrastive Learning.}}
\hspace{0.1cm} To better shape the cross-modal embedding space, we mine the correlations between training samples for superior inter-sample contrasts.

For inter-sample modeling, the key is to measure sample semantic similarity. While for TSG, category clusters are not intuitive. 
Thus, we consider using text queries as the bridge for correlation establishment, as language essentially refers to high-level semantics. We leverage the pre-trained Transformer Bert~\cite{b17} to extract query features and group samples into $K$ clusters $\Lambda_1, \Lambda_2, ..., \Lambda_K$. Note that the same sample could appear in different clusters. Within one batch of size $B$, we randomly select $N$ semantic clusters (B is divisible by N).

\noindent \underline{Event-Query Multi-Modal Contrast.}
We regard the events and queries from the same cluster as the positive pairs, while those from different clusters as negative pairs. Specifically, denoting the positive set of ${i}$-th sample as $\Psi^{\mathrm{+}}_{i}$, and the negative set as $\Psi^{\mathrm{-}}_{i}$, inter-sample multi-modal contrastive learning is:
\begin{align}
\mathcal{L}_{\mathrm{erml}} = & \sum_i - \log \frac{\sum_{m \in {\Psi^{\mathrm{+}}_{i}}}  \exp(\mathbf{v}_i^{\mathrm{ev}} \cdot \mathbf{q}_{m}/ \tau)}{\sum_{j \in \{ \Psi^{\mathrm{+}}_{i} \cup \, \Psi^{\mathrm{-}}_{i} \}
} \exp(\mathbf{v}_i^{\mathrm{ev}} \cdot \mathbf{q}_{j}/ \tau)}.
\end{align}
where $\tau$ is temperature coefficient, $\cdot$ is normalized dot product.

\noindent \underline{Vision Uni-Modal Contrast.} 
\hspace{0.1cm}
To model the cluster semantics among visual uni-modality, we construct the positive set $\Psi^{\mathrm{+}}_{\mathrm{i}}$ of $i$-th sample by joining events from the same cluster of $i$-th sample, while build the negative set $\Psi^{\mathrm{-}}_{\mathrm{i}}$ by leveraging event features from the other clusters. That is,
\begin{align}
\mathcal{L}_{\mathrm{erun}} = & \sum_i - \log \frac{\sum_{m \in {\Psi^{\mathrm{+}}_{\mathrm{i}}}}  \exp(\mathbf{v}_i^{\mathrm{ev}} \cdot \mathbf{v}_{m}^{\mathrm{ev}} / \tau)}{\sum_{j \in \{ \Psi^{\mathrm{+}}_{\mathrm{i}} \cup \, \Psi^{\mathrm{-}}_{\mathrm{i}} \}
} \exp(\mathbf{v}_i^{\mathrm{ev}} \cdot \mathbf{v}_{j}^{\mathrm{ev}}  / \tau)}.
\end{align}

\subsection{Explicit Stage: Unified Grounding}
By learning modality-alignment in the implicit stage, we obtain refined grounding pseudo-labels extended from annotated partial labels using the event detector (in Eq.~(\ref{eq:instance})). However, partial labels are unavailable during inference, making the event detector not work. Besides, lacking explicit grounding optimization causes somehow noise in pseudo-labels. 

To tackle these issues, we bridge another \textbf{explicit stage} after the implicit stage, by employing one off-the-shelf model from the fully-supervised research line. Specifically, we first treat the grounding pseudo-labels $[\widehat{s}, \, \widehat{e}]$ from the implicit stage as rough ground-truth for all samples in the training set, then optimize this fully-supervised model with explicit grounding objectives. 
At test time, we could solely utilize the fully-supervised model for direct inference. Compared with existing works~\cite{b18,b11} that require frame-by-frame matching during inference time, our framework directly outputs event timestamps, which is straightforward and easy to apply.

Although the contrast-unity framework is simple, the insight contained is non-trivial. \textit{Structurally}, this framework unites full-partial supervisions, enabling the partial setting to enjoy the superior grounding bonus from the existing fully-supervised methods. 
\textit{Functionally}, this framework bypasses the labeling gap between training and inference for partial supervision, by only utilizing the fully-supervised model for efficient inference. More importantly, our framework is flexible to handle supervisions ranging from single-frame to full annotation, enabling to jointly learn from wider data.

\subsection{Training and Inference}
\noindent \textbf{The implicit stage.} Partial labels provide limited yet precise supervision for the location of event mask $\mathbf{m}$ generated by predicted $[\widehat{s}, \, \widehat{e}]$, that is, the annotated short-clip is required to be included in the predicted event interval: 
\begin{equation}   \label{eq:partial}
{\mathcal{L}_{\mathrm{grnd}}=\mathrm{max}(t^e-\widehat{e}, \ \ \widehat{s}-t^s, \ 0),
}
\end{equation}
Above all, the implicit stage is optimized by balancing the contrastive loss $\mathcal{L}_{\mathrm{cont}}$ and grounding loss $\mathcal{L}_{\mathrm{grnd}}$ with ratio $\gamma$. 
We obtain acceptable $[\widehat{s}, \, \widehat{e}]$ grounding pseudo-labels this way.

\noindent \textbf{The explicit stage} treats pseudo-labels $[\widehat{s}, \, \widehat{e}]$ as the ground-truth labels to optimize another fully-supervised model~\cite{b19,b20,b21,b22,b23} (off-the-shelf), for grounding refinement. 
For efficient inference, we directly apply the fully-supervised model.

\section{Experiments and Results}
\begin{table}[t]
\footnotesize
\setlength\tabcolsep{0.8pt}
\centering
\caption{\textbf{Comparison with State-of-the-art.} F, W, SG and SC denotes Full, Weak, Single-frame and Short-Clip supervision, the last two is in the range of PTSG. Note that D3G and G2L uses C3D/VGG features on Charades-STA.}
\vspace{-0.3cm}
\begin{tabular}{cc|c|cccc|cccc}
\toprule 
\multicolumn{2}{c|}{\multirow{2}{*}{}} & \multirow{2}{*}{Method} & \multicolumn{4}{c|}{Charades-STA} & \multicolumn{4}{c}{ActivityNet Captions} \\ \cline{4-11} 
\multicolumn{2}{c|}{} &  & R@0.3 & R@0.5 & R@0.7 & mIoU & R@0.3 & R@0.5 & R@0.7 & mIoU \\ \hline
\multicolumn{2}{c|}{\multirow{7}{*}{F}} & TMLGA~\cite{b20}  & 69.62 & 50.11 & 32.50 & \multicolumn{1}{c|}{48.28} & 51.28 & 33.04 & 19.26 & 37.78 \\
\multicolumn{2}{c|}{} & LGI~\cite{b26}  & 72.96 & 59.46 & 35.48 & \multicolumn{1}{c|}{51.38} & 58.52 & 41.51 & 23.07 & 41.13 \\
\multicolumn{2}{c|}{} & SDN~\cite{b21}  & 73.71 & 59.89 & 41.80 & \multicolumn{1}{c|}{54.13} & 60.88 & 42.03 & 26.36 & 43.38 \\
\multicolumn{2}{c|}{} & CBLN~\cite{b27}  & -- & 61.13 & 38.22 & \multicolumn{1}{c|}{--} & 66.34 & 48.12 & 27.60 & -- \\
\multicolumn{2}{c|}{} & D-TSG~\cite{b28}  & -- & 65.05 & 42.77 & \multicolumn{1}{c|}{--} & -- & 54.29 & 33.64 & -- \\
\multicolumn{2}{c|}{} & G2L*~\cite{b23}  & -- & 47.91 & 28.42 & \multicolumn{1}{c|}{--} & 67.28 & 51.68 & 33.35 & 48.88 \\
\multicolumn{2}{c|}{} & BM-DETR~\cite{b22}  & 78.46 & 63.10 & 36.44 & \multicolumn{1}{c|}{56.42} & 67.33 & 50.23 & 30.88 & 48.23 \\

\hline \hline
\multicolumn{2}{c|}{\multirow{7}{*}{W}} & LCNet~\cite{b30}  & 59.60 & 39.19 & 18.87 & \multicolumn{1}{c|}{--} & 48.49 & 26.33 & -- & -- \\
\multicolumn{2}{c|}{} & VCA~\cite{b31} & 58.58 & 38.13 & 19.57 & \multicolumn{1}{c|}{--} & 50.45 & 31.00 & -- & -- \\
\multicolumn{2}{c|}{} & CRM~\cite{b32} & 53.66 & 34.76 & 16.37 & \multicolumn{1}{c|}{--}  & 55.26 & 32.19 & -- & -- \\
\multicolumn{2}{c|}{} & CNM~\cite{b24}  & 60.39 & 35.43 & 15.45 & \multicolumn{1}{c|}{--}  & 55.68 & 33.33 & -- & -- \\
\multicolumn{2}{c|}{} & CPL~\cite{b12} & 67.07 & 48.83 & 22.61 & \multicolumn{1}{c|}{43.71} & 55.28 & 30.61 & 12.32 & 36.82 \\
\multicolumn{2}{c|}{} & SCANet~\cite{b33} & 68.04 & 50.85 & 24.07 & \multicolumn{1}{c|}{--} & 56.07 & 31.52 & -- & -- \\
\multicolumn{2}{c|}{} & UGS~\cite{b34} & 69.16 & 52.18 & 23.94 & \multicolumn{1}{c|}{45.20} & 58.07 & 36.91 & -- & 41.02 \\

\hline \hline
\multicolumn{2}{c|}{\multirow{5}{*}{SG}} & LGI~\cite{b26} & 51.94 & 25.67 & 7.98 & \multicolumn{1}{c|}{30.83} & 9.34 & 4.11 & 1.31 & 7.82 \\
\multicolumn{2}{c|}{} & PS-VTG~\cite{b29} & 60.40 & 39.22 & 20.17 & \multicolumn{1}{c|}{39.77} & {59.71} & 39.59 & 21.98 & {41.49} \\
\multicolumn{2}{c|}{} & ViGA~\cite{b11} & 71.21 & 45.05 & 20.27 & \multicolumn{1}{c|}{44.57} & {59.61} & {35.79} & {16.96} & {40.12} \\
\multicolumn{2}{c|}{} & D3G*~\cite{b18} & -- & 41.05 & 19.60 & \multicolumn{1}{c|}{--} & {58.25} & 36.68 & 18.54 & {--} \\
\multicolumn{2}{c|}{} & \textbf{Ours} & {\bf 75.09} & {\bf 61.51} & {\bf 32.69} & \multicolumn{1}{c|}{{\bf 52.31}} & {\bf 60.85} & {\bf 40.60} & {\bf 22.75} & {\bf 42.08} \\  

\hline
\multicolumn{2}{c|}{\multirow{2}{*}{SC}} & \textbf{Ours (2s)} & {\bf 75.33} & {\bf 62.49} & {\bf 33.71} & \multicolumn{1}{c|}{{\bf 53.24}} & {\bf 61.14} & {\bf 43.51} & {\bf 25.79} & {\bf 43.67} \\ 
\multicolumn{2}{c|}{} & \textbf{Ours (4s)} & {\bf 76.83} & {\bf 62.51} & {\bf 35.27} & \multicolumn{1}{c|}{{\bf 54.98}} & {\bf 64.22} & {\bf 45.89} & {\bf 27.38} & {\bf 45.70} \\ 
\bottomrule
\end{tabular}
\vspace{-0.4cm}
\label{tab:SOTA}
\end{table}

\subsection{Datasets and Evaluation Metrics}
\noindent \textbf{Datasets.} We use the most common benchmarks in the TSG task: Charades-STA, ActivityNet Captions and Charades-CD.

\noindent \textbf{Evaluation Metrics.} Following the existing works~\cite{b11,b12}, we evaluate through ``R@K, IoU=M'', \textit{i.e.}, the percentage of predicted moments with Intersection over Union (IoU) greater than M in the top-K recall. To simplify the notation, we note ``R@1, IoU=M'' as ``R@M'' in the following sections.
\begin{table}[t]
\footnotesize
\setlength\tabcolsep{4pt}
\centering
\caption{\footnotesize \textbf{Ablation study of quadruple constraint pipeline.} We evaluate the quality of pseudo-labels under single-frame annotations.
All losses jointly contribute to the best performance.}
\vspace{-0.3cm}
\begin{tabular}{c|cccc|ccc}
\toprule 
& $\mathcal{L}_{\mathrm{raml}}$ & $\mathcal{L}_{\mathrm{raun}}$ & $\mathcal{L}_{\mathrm{erml}}$ & $\mathcal{L}_{\mathrm{erun}}$ & R@0.5 & R@0.7 & mIoU \\ 
\hline \hline
A1 & \checkmark & &  &  & 31.78 & 8.91 & 44.30 \\
A2 & \checkmark & \checkmark &  &  & 36.41 & 10.63 & 46.41 \\
A3 & \checkmark &  & \checkmark &  & 62.99 & 25.90 & 57.67 \\
A4 & \checkmark & \checkmark & \checkmark &  & 67.64 & 27.52 & 58.90 \\
A5 & \checkmark & & \checkmark & \checkmark & 67.40 & 27.58 & 58.71 \\
A6 & \checkmark & \checkmark & \checkmark & \checkmark & 71.32 & 28.59 & 60.50 \\ 
\bottomrule
\end{tabular}
\vspace{-0.4cm}
\label{tab:ablation}
\end{table}

\begin{table}[t]
\footnotesize
\setlength\tabcolsep{4pt}
\centering
\caption{\textbf{Effectiveness of partial annotations.} For single-frame or short-clip labels from various types of labeling distributions, the implicit stage demonstrates strong robustness and superiority for generating high-quality pseudo-labels.}
\vspace{-0.3cm}
\begin{tabular}{c|c|cccc}
\toprule
Setting & Distribution & R@0.3 & R@0.5 & R@0.7 & mIoU \\ \hline  \hline
\multirow{3}{*}{\begin{tabular}[c]{@{}c@{}}Single\\ Frame\end{tabular}}
 & Uniform-1 & 97.77 & 71.32 & 28.59 & 60.50 \\
 & Uniform-2 & 97.25 & 71.50 & 28.94 & 60.35 \\
 & Uniform-3 & 97.71 & 72.04 & 29.48 & 60.69 \\ \cline{2-6} 
 & Gaussian & 98.23 & 78.08 & 34.70 & 63.17 \\  \hline \hline
\multirow{3}{*}{\begin{tabular}[c]{@{}c@{}}Short\\ Clip\end{tabular}} 
 & 2-seconds & 99.69 & 89.05 & 40.46 & 67.01 \\
 & 3-seconds & 99.84 & 89.94 & 44.52 & 68.07 \\
 & 4-seconds & 99.89 & 92.79 & 52.76 & 70.56 \\
\bottomrule
\end{tabular}
\vspace{-0.3cm}
\label{tab:singleframe}
\end{table}

\begin{table}[t]
\footnotesize
\setlength\tabcolsep{4pt}
\centering
\caption{\textbf{framework Generalization.} Bridging our partial branch to various fully-supervised methods brings promising results.}
\vspace{-0.3cm}
\begin{tabular}{c|c|cccc}
\toprule 
Method & Label & R@0.3 & R@0.5 & R@0.7 & mIoU \\ \hline  \hline
\multirow{3}{*}{\begin{tabular}[c]{@{}c@{}}IA-Net \cite{b19}\end{tabular}} & Full & 68.87 & 57.00 & 28.27 & 46.63 \\
 & Single-Frame & 65.36 & 53.60 & 25.87 & 44.70 \\
 & Short-Clip & 67.22 & 55.52 & 27.41 & 45.74 \\\hline \hline
\multirow{3}{*}{\begin{tabular}[c]{@{}c@{}}TMLGA \cite{b20}\end{tabular}} & Full & 69.62 & 50.11 & 32.50 & 48.28 \\
 & Single-Frame & 67.61 & 45.08 & 26.34 & 45.36 \\
 & Short-Clip & 67.26 & 50.97 & 28.55 & 46.10 \\\hline \hline
\multirow{3}{*}{\begin{tabular}[c]{@{}c@{}}SDN \cite{b21}\end{tabular}}
& Full & 73.71 & 59.89 & 41.80 & 54.13 \\
 & Single-Frame & 71.57 & 54.66 & 28.34 & 48.65 \\ 
 & Short-Clip & 72.09 & 56.43 & 32.08 & 49.91 \\ \hline \hline
 \multirow{3}{*}{\begin{tabular}[c]{@{}c@{}}BM-DETR \cite{b22}\end{tabular}}
& Full & 78.46 & 63.10 & 36.44 & 56.42 \\
 & Single-Frame & 75.09 & 61.51 & 32.69 & 52.31 \\ 
 & Short-Clip & 75.33 & 62.51 & 33.71 & 53.24 \\ 
\bottomrule
\end{tabular}
\vspace{-0.4cm}
\label{tab:fullsuoervision}
\end{table}

\begin{table}[t]
\footnotesize
\setlength\tabcolsep{4pt}
\centering
\caption{\footnotesize \textbf{Contrast designs.} In inter-sample modeling, our relevance mining uses similar queries brings great gains, comparing to vanilla data augmentation. `Pos' and `Neg' refer to mining for positive samples and negative samples, respectively.}
\vspace{-0.3cm}
\begin{tabular}{c|c|cc|cc}
\toprule 
\multirow{2}{*}{} & \multirow{2}{*}{Representation} & \multicolumn{2}{c|}{Inter-Sample} & \multirow{2}{*}{R@0.7} & \multirow{2}{*}{mIoU} \\ \cline{3-4}
 &  & Relevance & Sample &  &    \\ \hline \hline
B1 & \multirow{3}{*}{(event, query)} & Augment & Pos+Neg & 20.81 & 54.02  \\
B2 &  & Similar & Pos & 3.06 &  35.32  \\
B3 &  & Similar & Pos+Neg & 28.59 & 60.50   \\   \hline
B4 &  (short-clip, query) & Similar & Pos+Neg & 21.01 &  56.11  \\
B5 & (video, query) & Similar & Pos+Neg & 18.67 & 51.39  \\ 
\bottomrule
\end{tabular}
\vspace{-0.4cm}
\label{tab:representation}
\end{table}

\subsection{Comparison with State-of-the-art}

\noindent \textbf{Single-frame supervision.}
Table~\ref{tab:SOTA} demonstrates comparisons across multiple IoU thresholds on both datasets. For the sake of fairness, we use the identical single-frame annotations following~\cite{b11}.
Our framework achieves new state-of-the-art under all IoU regimes, by a large margin. For example, 7.74\% mIoU gains over the previous SOTA on Charades-STA. Moreover, our method gains more on the rigorous evaluation than loose regimes, \textit{e.g.}, 3.88\% gains for R@0.3 \textit{vs.} 12.42\% gains for R@0.7, comparing to ViGA~\cite{b11}. Despite being single-frame supervision, our method is even comparable with some earlier fully-supervised methods~\cite{b25,b26}. We also offer results on Charades-CD dataset to test OOD data generalization ability, which also surpass existing methods by a large margin.

Note that, ActivityNet Captions is challenging even for fully-supervised methods, with only 1-7\% gaps between full-weak supervisions. Such small gaps somehow limit our potentiality. Still, we achieve the state-of-the-art performance on all regimes. With more advanced fully-supervised methods becoming available, our results can be further improved.

\noindent \textbf{Short-clip supervision.}
Table~\ref{tab:SOTA} experiments on short-clip of 2/4 seconds. A steady improvement could be witnessed with longer annotation intervals, further narrowing the PTSG-FTSG performance gap: only 1.5\% gap over FTSG SOTA for 4s.

\subsection{Ablation Study \& Discussion}
We conduct thorough ablations to dissect all key components, using single-frame annotations on Charades-STA.

\noindent \textbf{Framework Robustness.} 
Partial labels possess a high degree of freedom in event intervals, bringing great challenges to framework robustness. Table~\ref{tab:singleframe} simulates pseudo-label quality in implicit stage with multiple annotation samplings for different distributions/durations. Our framework shows consistent effectiveness to various annotations, proving strong robustness.

\noindent \textbf{Contribution of Quadruple Contrasts.}
To achieve great representations, we design quadruple contrasts: for intra-sample, uni-modal loss $\mathcal{L}_{\mathrm{raun}}$ and multi-modal loss $\mathcal{L}_{\mathrm{raml}}$; for inter-sample, uni-modal loss $\mathcal{L}_{\mathrm{erun}}$ and multi-modal loss $\mathcal{L}_{\mathrm{erml}}$.
In Table~\ref{tab:ablation}, the single $\mathcal{L}_{\mathrm{raml}}$ (A1) causes poor pseudo-labels for the partial branch. A2 adds $\mathcal{L}_{\mathrm{raun}}$ to encourage event-back separation, thus obtaining clear improvements.
Happily, inter-sample modeling brings immediate gains. By introducing $\mathcal{L}_{\mathrm{erml}}$ to A1, A3 gets more than 13\% mIoU gains; by adding $\mathcal{L}_{\mathrm{erun}}$ to A3, A5 further gets 1.1\% mIoU gains, showing the advantages of sample relationship modeling. In conclusion, all losses are essential and jointly contribute to the best results.

\noindent \textbf{Effectiveness of Representation Learning.}
We propose (event, query) aligned pairs over (video, query) or (short-clip, query) pairs. As shown in Table~\ref{tab:representation}, event-query representation achieves a 4.4\% mIoU gain over video-query. The short-clip-query approach, limited to partial clips, hinders grounding completeness and underperforms.

\noindent \textbf{Effectiveness of Inter-Sample Contrasts.}
\hspace{0.1cm} To obtain sample relationships by clustering, Table~\ref{tab:representation} uses a baseline: randomly augment video as positive. Comparing B3 to B1, we find query-based semantic consistency effective for videos and hard sample mining more efficient than simple augmentation. Additionally, B3 vs. B2 highlights the effectiveness of inter-video negative samples during training.

\noindent \textbf{Generalization of The Explicit Stage.}
Our method bridges the gap between PTSG and FTSG, thus can process data from various supervisions. And Table~\ref{tab:fullsuoervision} evaluates its generalization, by employing three typical fully-supervised methods (IA-Net~\cite{b19}, TMLGA~\cite{b20}, SDN~\cite{b21}, BM-DETR~\cite{b22}) in the explicit stage. Despite varying annotations, our PTSG method performs comparably to fully-supervised approaches. Future advancements promise further improvement.

\section{Conclusion}
We propose partial supervision for TSG to balance performance and annotation effort. Our novel contrast-unity framework employs a two-stage approach: implicit-explicit progressive grounding. In the implicit stage, quadruple contrastive learning aligns event-query representations, generating high-quality pseudo-labels. These pseudo-labels are then used in the explicit stage to train a fully-supervised model for refined grounding. Experiments demonstrate our framework's superior performance.

\section{Acknowledgements}
This work is supported by STCSM (No. 22511106101), 111 plan (No. BP0719010), and State Key Laboratory of UHD Video and Audio Production and Presentation.

\end{document}